# Towards Meaningful Transparency in Civic AI Systems

Dave Murray-Rust, Human Centred Design, TU Delft
Kars Alfrink, Sustainable Design Engineering, TU Delft
Cristina Zaga, Human Centred Design, University of Twente

Abstract
Artificial intelligence has become a part of the provision of governmental services, from making decisions about benefits to issuing fines for parking violations. However, AI systems rarely live up to the promise of neutral optimisation, creating biased or incorrect outputs and reducing the agency of both citizens and civic workers to shape the way decisions are made. Transparency is a principle that can both help subjects understand decisions made about them and shape the processes behind those decisions. However, transparency as practiced around AI systems tends to focus on the production of technical objects that represent algorithmic aspects of decision making. These are often difficult for publics to understand, do not connect to potential for action, and do not give insight into the wider socio-material context of decision making. In this paper, we build on existing approaches that take a human-centric view on AI transparency, combined with a socio-technical systems view, to develop the concept of *meaningful transparency* for civic AI systems: transparencies that allow publics to engage with AI systems that affect their lives, connecting understanding with potential for action.

## 1. Introduction

There is a strong and intensifying trend of integrating artificial intelligence (AI) and machine learning (ML) technologies into public services such as healthcare, education, emergency services, and transportation. Automated decision-making (ADR) systems create outcomes for residents, from everyday questions of parking violations through automatic number plate recognition to larger questions of parole, welfare, etc. This is often part of government agendas around technology use, for example, the digital agendas of Dutch local councils (*NL DIGIbeter*, 2020) that emphasize the potential of digital technology to make things better. The underlying narrative is often that replacing human judgment with computational systems will lead to more efficiency and fairer outcomes.

However, despite promises of impartiality or fairness, these technologies often operate in ways that are incorrect and unfair (Barocas & Selbst, 2016; Deck et al., 2024; Friedler et al., 2021; Pasquale, 2015) and do not live up to their promised levels of functionality (Raji et al., 2022). The AI Now 2025 Landscape report identifies five key takeaways, of which two are key to our argument: AI solutions displace grounded expertise around entrenched social problems, and the use of AI technology is frequently coercive, violating rights and undermining due process (Brennan et al., 2025, p. 47).

A particular challenge in this situation is that decision subjects—people who are subjected to AI decisions (Gemalmaz & Yin, 2022)—lose their agency in the process, and cannot influence AI-driven processes. While bureaucratic decisions are complex, they are also comprehensible to humans, so there is the potential for redress of unfair or incorrect decisions. Automated decisions are less comprehensible to lay publics: the workings of the system are not clear, making it challenging for subjects to assert their rights (Brennan et al., 2025, p. 66). They are also difficult to understand for experts, as they typically involve technology commissioned from private organisations, datasets collected in various ways; a combination of secrecy and opacity limits public understanding of system operation (Burrell, 2016). Even when responsible AI methods (Dignum, 2017; Lu et al., 2022) are deployed, it is difficult to create fair and equitable systems without deep participation (Birhane et al., 2022a; Guo et al., 2025).

Addressing the opacity of AI systems requires clarifying their functions and providing a means for redress. The FAccT community (*ACM FAccT*, n.d.) looks at fairness, accountability, and transparency as key principles for making more just AI systems. While AI transparency has been an active area of research, much of the attention focuses on explainability (XAI) or interpretability as mechanisms to enable transparency: making models either simple enough that they can be understood by humans, or able to provide meaningful accounts of their own operation (Barredo Arrieta et al., 2020).

However, despite movements towards more human-centred ideas of transparency that connect explanations to their purposes (Ehsan et al., 2022; Ehsan & Riedl, 2020), most AI explanation work remains *technical*, in that it is difficult for lay users to understand; *narrow*, in that it covers the internal operation of models, not their whole sociotechnical context; and *disconnected* from the possibility to make change. The question of *transparency* within the AI systems community is often disconnected from previous critical work on transparency (Corbett & Denton, 2023a). It fails to achieve the desired accountability and systemic change (Ananny & Crawford, 2016).

Taking a step back from engineering oriented approaches, we situate this work in the broader field of transparency research (Albu & Flyverbom, 2019; Flyverbom, 2022; H. K. Hansen et al., 2015; Hood & Heald, 2012) which intimately connects transparency with the aims it seeks to support, typically better governance, reduced corruption, greater equitability and so on. We draw on this work, including approaches that see transparency as an "informational mechanism necessary for performing the virtues of truthfulness, justice, and prudence" (das Neves & Vaccaro, 2013) but more specifically as a meta-ethical principle (Turilli & Floridi, 2009) that enables accountability in support of safety, welfare, fairness and other socially desirable qualities. On an individual level, transparency can enhance understanding of decision-making processes and provide a basis for contesting those decisions. On a societal level, it enables organizations to be held accountable and encourages better behavior.

In this paper, we reframe the current perspectives on transparency for AI in public services using a relational lens to construct the concept of *meaningful transparency* as situated within a range of social structures, connected to decision subjects, and able to make change. The term 'meaningful transparency' has been used before around the operation of AI and IoT systems (Cobbe et al., 2021; Norval et al., 2021, 2022; Norval & Singh, 2024; B. G. Schor et al., 2022), as well as appearing ad hoc from participants in interviews about what transparencies they would like in emotion recognition (Grill & Andalibi, 2022) or content moderation systems (Suzor et al., 2019). Here, we construct the term in the context of civic AI: interleaved physical and digital infrastructure in cities making use of AI and ML for decision-making across social services, infrastructure, public safety, and more (Madaio & Martin, 2020). We look at what transparencies are meaningful when diverse and distributed publics engage with services that increasingly use AI systems in their operation and decision-making.

We make three contributions in this paper:
1. We bring together threads from transparency literature with AI transparency practice to develop relational ideas of transparency in AI systems, in particular, *lived transparencies* as those that connect to the lifeworld of individuals.
2. We build a conceptual framework for looking at multiple transparencies, and develop a definition of a *meaningful transparency* as one that can be technically produced, connects to the lifeworld of individuals, and supports the possibility for action.
3. We discuss current and emerging practices that could support meaningful transparency and lay the groundwork for future work.

Together, this creates a solid definition of meaningful transparency as a relational, socially situated approach to transparency, where the technical production of transparency artefacts is aligned with citizens' needs and connects to possibilities for them to make change. We start with a survey of the foundations of transparency and related work (Section 2) and develop the context of civic AI (Section 3). Based on this, we develop *lived transparencies* and *meaningful transparency* concepts (Section 4) before exploring emerging practices that can help to create them (Section 5).

## 2 Background and Related Work

To establish a conceptual foundation for rethinking transparency in Civic AI, we combine three areas of focus: the foundations of modern transparency as a tool for better governance; the construction of transparency in online social settings; and current trends in AI and HCI that directly work with algorithmic transparency in action. Two background ideas frame our survey of transparency: firstly, that it "is an ideal rather than an end state ... ephemeral and always in the making", and secondly, the fact that there are enough ways to describe transparency that it becomes more a family resemblance than a precisely formalisable term (H. K. Hansen et al., 2015). By casting a broad net over what transparency is, we bring together concepts that help nuance and situate our ideas about what meaningful transparency should be.

### 2.1 Foundations of Transparency

Prior to discussion around technology, the prime discourse around transparency was related to the operation of governments and public organisations (Heald, 2012) or corporations (das Neves & Vaccaro, 2013). Transparency is seen as way to operationalise ethics around information systems (Vaccaro & Madsen, 2009) a way to support accountability, safety, welfare and informed consent, a virtuous practice to "truthfulness, justice, and prudence" (das Neves & Vaccaro, 2013).

Heald (2012) sets out key properties that transparency may have. Transparency may operate in different directions: upwards, where superiors observe the behaviour of subordinates; downwards, when subjects see into the actions of decision makers; outwards, as organisations gain a view outside; and inward, where organisations make themselves transparent to those outside. Meijer (2014) adds another direction: inverse, where transparency removes some need for accountability. It may be concerned with events (or objects), or with the processes that connect and generate those events. Transparency may be produced retrospectively—for example, quarterly reports—or in real-time and continuously available (Heald, 2012). Meijer (2014) also adds a typology of three key kinds of transparency: virtue-based transparency is a normative system of actors being open about their goals and considerations; relational transparency is opening up inner workings for the benefit of particular observers; and systemic transparency involves looking across sets of these relations. Overall, while access to truth and the associated knowledge that comes with it is important, "expected outcomes [of transparency] range from increased visibility and efficiency to accountability, authenticity, participation, involvement, empowerment, emancipation, and trust" (ibid.).

This last part connects to Heald's final distinction between nominal transparency and effective transparency. Transparency is more than the provision of information; if actors cannot understand and make use of the information provided, it is nominal transparency, without effective force. Hence, there is a need to "embrace simplicity and comprehensibility" in the production of transparency (Larsson, 1998). In contrast to what could be considered a factual property of 'openness', it is "possible for an organization to be open about its documents and procedures yet not be transparent to relevant audiences" as it lacks the "external receptors capable of processing the information made available" for effective transparency (Heald, 2012). Stohl (2016) characterises this as the difference between visibility—a fact of information accessibility—and transparency, which is the ability of others to engage with the visible information. This allows for accidental or strategic opacities to be created, for instance by such large scale release of data that it is impossible to find the items of interest. Even with good intention "Unless the government requires disclosure in forms that the public finds digestible, the information is often released in ways that provide little de facto transparency." (Etzioni, 2016).

### 2.1.1 Organisational Transparency

This question of how transparency affects the operations of organisations is brought together and critiqued under the name 'organisational transparency' by Flyverbom and others (Albu & Flyverbom, 2019; Flyverbom, 2019, 2022). Albu and Flyverbom (2019) develop the conceptualisations, conditions and consequences of organizational transparency, and make a key distinction between two kinds of transparency:
- Verification approaches allow one to see 'what the organisation is really like', assess legitimacy and take action
- Performative approaches (note - performative is not used in a perjorative sense here, rather in the way that transparency is performed as a social process) look at how transparencies "produce a different world" (Loxley, 2007, p.2). This highlights the generative nature of transparency projects, and the ways that they modify the organisations they render visible.

This work builds the foundation for a treatment of transparency as a socio-material practice, connecting the production of transparency to its understanding and eventual effects in the world.

### 2.1.2 Transparency in and around digital systems

Shifting slightly from organisational work, to look at the foundations of practical design for transparency in information systems or Human-Computer Interaction work, one of the key threads is social translucency (Erickson & Kellogg, 2000). This looks at designing social digital systems - typically in the context small group settings - from a point of view of who can see which information at which point, negotiating the balance between privacy and transparency. Here, transparency is viewed from the point of view of rendering actions visible to others, as the basis for coherent activity. This way of thinking about transparency highlights not just the functional ideas of visibility, but the relational possibilities built on a substrate of activity (McDonald et al., 2012). Of interest here is the idea that to support social translucence, beyond simple visibility, systems should support the awareness that is necessary both for following and understanding social rules, and creating accountability for actions (Erickson & Kellogg, 2000). This is picked up to create the idea of 'social transparency' as a result of information exchange. Stuart et al. (2012) disentangle three kinds of transparency: identity transparency, or who is exchanging information; content transparency, or what has happened to the information; and interaction transparency, or the possibility of observing actions taken within interactions. In these situations, the primary focus is on horizontal and downwards transparencies (Heald, 2012) - transparency about the action of one's peers, used to construct the social fabric, and the question of what is surveilled from above by system creators. Despite the surface simplicity, there is a deep philosophical import to the idea that transparency is not a given in these systems; visibility is not a natural function of the world, but rather a field whose rules must be purposefully constructed to create desirable systems. The ability to carry out coherent action has been a site of contestation in algorithmic systems. Amazon's Mechanical Turk purposefully created an atomised workforce, where individual 'Turkers' were not able to share information, in particular about the jobs they were commissioned to carry out - reducing both upwards and horizontal transparency. In response Turkopticon (L. C. Irani & Silberman, 2013) was created as a piece of infrastructure that rendered visible the actions of the commissioners (upwards transparency), especially those with poor behaviours towards Turkers. The communication between workers (horizontal transparency) allowed them to collectively act towards better quality jobs, and also created a "web of relationships of common cause", going beyond the technology itself (L. Irani & Silberman, 2014) .

Another, more subtle take on what transparency means in relation to HCI is the distinction between 'seeing into' and 'seeing through' (Christensen & Cheney, 2015). Where one use of the word is reducing or eliminating obstacles to information visibility, in HCI studies, ''transparency'' is more likely to refer to a condition of information invisibility, such as when an application or computational process is said to be transparent to the user (Turilli & Floridi, 2009). Andrada et. al (2023) separate these two approaches as reflective transparency, that allows "greater oversight, accountability, ability to intervene, or simply understanding", and transparency in use, where the interface disappears as attention is on the task being performed - "seeing through a technology to competently and fluently act while using it". A related view is seeing across a system (Ananny & Crawford, 2016), where the seeing is not just into the workings of a system, but its constitution as a complex actor-network, with relational significance.

Together, these paint a picture of transparency as a relational concept, that holds between particular organisations and groups, with the production of transparency forming the possibilities for action, or an ordering principle (Flyverbom, 2015), rather than being purely about holding organisations to account. Or alternatively: "how [transparency] is producing, for example new regimes of visibility, comparability and control." (Christensen & Cheney, 2015).

## 2.2 AI transparency in HCI

HCI research on AI transparency encompasses approaches ranging from technical explainability to more situated, relational, and contextual perspectives. The field includes work that positions HCI as contributing to explainable AI through user-centered design and empowerment (Adadi & Berrada, 2018), while other research develops critical perspectives that distinguish transparency from explainability and emphasize broader social and political dimensions (Corbett & Denton, 2023b). More contextually-oriented work examines how power asymmetries between platforms and users can undermine transparency mechanisms, arguing for situated rather than universal interventions that account for specific social and political complexities (Ramesh et al., 2022).

HCI uses empirical methods to study how users actually experience transparency mechanisms. Studies examine user perceptions of algorithmic opacity and identify factors that influence attitudes toward transparency and subsequent behavior change (Eslami et al., 2019). Experimental work tests how different amounts and types of algorithmic information affect user trust and decision-making, comparing explanation approaches to determine which transparency functions—awareness, correctness, interpretability, and accountability—each serves (Kizilcec, 2016; Rader et al., 2018).

HCI research produces frameworks and tools for implementing transparency in AI systems. Researchers have developed standardized documentation approaches based on practices from other fields, creating structured formats for communicating model information to various audiences (Arnold et al., 2019; Mitchell et al., 2019). Other work provides systematic frameworks for integrating transparency requirements into development processes and models for designing transparency that accounts for how users process information (Felzmann et al., 2020; Liao & Sundar, 2022). Research identifies gaps between stakeholder needs and current transparency approaches, emphasizing the need to balance statistical practices with social practices for ensuring fairness in high-stakes public sector applications (Veale et al., 2018). Interestingly, it appears that emerging AI regulation may create greater scope for transparency than existing FoI regulation (Olsen et al., 2024), hinting at new possibilities for governance.

Together, these empirical, critical, and design-oriented approaches demonstrate HCI's distinctive contribution to AI transparency: grounding technical solutions in human realities while remaining attentive to the social and political contexts in which transparency operates. This multifaceted approach positions HCI to address not just how to make AI systems transparent, but whether and when transparency serves its intended purposes.

## 2.3 Limits of transparency

The transparency ideal rests on fundamentally flawed assumptions about the relationship between visibility and understanding. Transparency falsely equates seeing with knowing, assumes systems can be held static for observation, and treats viewers as subjects rather than objects of power (Ananny & Crawford, 2016) . The explanatory process itself is inherently social, contrastive, and biased, with people employing cognitive shortcuts and social expectations rather than objective analysis (Miller, 2019). Moreover, AI systems possess inherent technical obscurity that cannot be eliminated through design choices (Brauneis & Goodman, 2017; Felzmann et al., 2019), compounded by obscurity around information such as "the public purpose for which the algorithm was developed, the contract terms that govern data ownership and access, and plans for validation and follow-up" (Brauneis &

Goodman, 2017, p. 131). Current explainable AI approaches compound these problems by focusing narrowly on algorithmic mechanisms while ignoring crucial socio-organizational contexts (Ehsan et al., 2021).

Transparency requirements vary dramatically across stakeholders, domains, temporal contexts, and implementation settings, making universal approaches ineffective. Different user groups hold conflicting expectations for the same transparency interventions, and contextual factors significantly influence effectiveness (Felzmann et al., 2019). Domain-specific challenges emerge across healthcare, smart cities, and public services, each requiring tailored approaches that account for professional workflows and multidisciplinary team structures (Brauneis & Goodman, 2017; Brown et al., 2019; B. G. S. Schor et al., 2024a). Users approach transparency with different mindsets—utilitarian, interpretive, or critical—that shape their information needs and interpretation processes (Buschek et al., 2022). Furthermore, user engagement patterns influence reactions to transparency interventions, while temporal changes in systems create ongoing challenges for maintaining meaningful transparency (Ananny & Crawford, 2016; B. G. S. Schor et al., 2024a; West et al., 2025).

Academic transparency research frequently fails to translate into effective real-world implementations. Most explainable AI work relies on researcher intuitions rather than empirical evidence from social sciences, missing the interactive and goal-oriented nature of human explanation (Miller, 2019). Experimental studies demonstrate that intuitively interpretable designs do not produce measurable behavioral benefits, with clear models failing to improve user decision-making (Poursabzi-Sangdeh et al., 2021). Users' expectations about AI functionality often diverge significantly from actual system behavior (Felzmann et al., 2019), while legal frameworks prove insufficient for achieving meaningful transparency (West et al., 2025). Research recommendations developed in academic settings frequently cannot be operationalized in clinical and other real-world contexts (B. G. S. Schor et al., 2024a), and technical approaches fail to address fundamental user questions and concerns (Buschek et al., 2022).

Transparency interventions often produce outcomes opposite to their intended effects, causing harm rather than enabling accountability. Visibility can disempower those being observed, strengthen untrustworthy actors, and create false confidence in systems too complex for human comprehension (Ananny & Crawford, 2016). Information overload from transparent systems impairs users' ability to detect errors and make appropriate corrections, while algorithmic formalism abstracts away social concerns that transparency should illuminate(Ehsan et al., 2021; Poursabzi-Sangdeh et al., 2021). Direct exposure to authentic AI models generates negative emotional reactions and long-term behavioral changes that suggest transparency can be fundamentally disruptive to user experience (West et al., 2025).

These limitations collectively suggest that transparency, as currently conceived, is insufficient for algorithmic accountability and that alternative governance mechanisms must work constructively with rather than against these inherent constraints.

## 2.4 Seeds of Meaningful Transparency

We frame this paper around the term "meaningful transparency" in the context of AI systems, but as is often the case, there are prior uses of this term, many of which we are in alignment with. Similarly, as we have traced some relational conceptualisations of transparency, there are calls that AI transparency should be handled in a relational manner.

Felzman et al. (2019) suggest that we should "understand transparency relationally, where information provision is conceptualized as communication between technology providers and users". This builds concretely on a typology from Weller (2019) that discusses various audiences for transparency outputs: developers; users; society at large; regulators, auditors or overseers; along with reasons for wanting the transparency: a general sense of why the system is working as it is; understanding a particular output; maintaining compliance and so on. This already sets the stage for transparency that is tailored to particular ends and audiences. *Relational transparency* is picked up by van Nunen et al. (2020; 2022) to capture the context, justifications, interpretability and other transparency relative to a audiences. In the context of intersectional algorithmic harms from civic AI, it is necessary to provide: "the kinds of explanations that users from different backgrounds require" (Van Nuenen et al., 2022), with co-design and related practices as an enabling methodology.

For the term *meaningful transparency*, we trace three separate evolutionary arcs. Firstly, Suzor et al. develop the term *meaningful transparency* in the context of content moderation, in order to explain what actions caused moderation, why the decision was taken, bias in policies and undue algorithmic influence (Suzor et al., 2019). It requires a deep engagement with the systemic complexities of the situations, and and should be "conceptualized as a communicative process of rendering account to the many stakeholders implicated in them". Of particular interest here is the idea that while single cases may be explained, large scale issues of biased decisions need a different form of transparency, rendered for a different audience, with "large-scale access to data on individual moderation decisions as well as deep qualitative analyses of the automated and human processes that platforms deploy internally." Secondly, in the context of emotion recognition, Grill and Andalibi (2022) coined *meaningful transparency* as a particular theme distilled from interviews with users, where they look at transparency in practice. In particular, these transparencies would "enable them to be more thoughtful about their behavior online", and "enable them to contest accuracy claims". The third, and most conceptually developed thread starts from Cobbe et al. (2021) and Norval et al. (Norval et al., 2021) where meaningful transparency is "targeted, useful transparency, providing information that is (i) relevant to the accountability relationships involved, (ii) accurate, complete, and representative, (iii) proportionate to the level of transparency likely to be required, and (iv) comprehensible by the relevant forums" (Cobbe et al., 2021). This fits with a call that meaningful processes require holistic views, "from commissioning of the system; through design of the model, selection of training data, and training and testing procedures; to making individual decisions; and on to the effects of those decisions and any subsequent investigations." This leads to the idea that disclosure should happen through "user-centric interfaces [...] designed around the needs of their intended recipients", and is explored in work with clinicians (B. G. S. Schor et al., 2024b) and IoT devices (Norval & Singh, 2024)

Despite the uses of the term, there is space for a theoretical construction of what at term like this means; in particular, one which grounds the meaningfulness of transparency not solely in properties of the information that carries the transparency, but in a relational construction between the information, its audience and the socio-technical systems in which it is embedded.

## 3 Context: Civic AI

Civic AI results from interleaving physical and digital infrastructure in cities, with increasing reliance on artificial intelligence methods and techniques as part of the digital layer. It also includes using predictive machine learning models derived from historical data to inform municipal decision-making across social services, infrastructure, public safety, and more (Madaio & Martin, 2020). Civic AI can be conceptualized as AI that allows for civic participation, though such participation can be either meaningful or tokenistic. To be meaningful, people's lived experiences must be brought to bear on AI systems design. An even stronger form of civic participation seeks to empower marginalized communities, requiring mechanisms to investigate, contest, influence, or even dismantle AI systems (Sieber et al., 2025). Civic AI has also been conceptualized as a tool for creative ideation and civic engagement in smart cities, where the hope is that such tools allow for a greater degree of inclusivity and flexibility than traditional approaches (Chauncey & McKenna, 2024).

Civic empowerment concerning AI can happen along at least three dimensions: institutional transparency about current and new AI-enabled government services so that civil society can participate in public debate about them, governments building AI systems that support citizens in daily tasks (i.e., "functional empowerment"), or actual civic participation in the design of public-sector AI systems (i.e., "democratic empowerment"). All these forms of empowerment face barriers, including cultures that privilege expertise and are risk-averse (Drobotowicz et al., 2023). Meaningful participation by citizens in civic AI systems design also requires they have the necessary digital and data literacy skills to do so (Madaio & Martin, 2020).

Between governments and individual citizens, public institutions such as libraries can play an important role as trusted intermediaries. They can offer neutral spaces for inclusive civic engagement in public sector AI initiatives, going beyond the awareness-raising and public education that is the current norm (Huang et al., 2024, 2025). Practitioners in human-centered explainable AI (HCXAI) consider participatory approaches to XAI, specifically in civic AI applications, a "new frontier." They emphasize that "who opens" the black box is as important as how it is opened (Ehsan et al., 2025).

An appeal of AI for civic participation is that of scale and synthesis. AI systems can include many voices and synthesize large volumes of public-contributed data. However, this scalability, often the product of AI systems' transnational nature, can produce tensions with local community participation needs (Sieber et al., 2025). Speculative design approaches have explored how civic AI could take the form of machines directly participating in civic governance (Wong, 2018).

Civic participation in AI faces a range of challenges. The unique properties of AI as a technology (opacity, instability, feedback-ladenness) can complicate traditional approaches to participation. AI models may be biased, which can then propagate throughout the systems they are embedded in, leading to harm to citizens. Integrating AI in civic systems leads to increased opacity, which can impede the accountability of city governments to their citizens (Madaio & Martin, 2020). There is a risk of "calculated publics" where communities' self-understanding becomes predominantly shaped by algorithmic viewpoints. Much of the literature on civic participation in AI is reluctant to explicitly grapple with the problem of political power (Sieber et al., 2025). Smart city initiatives can lead to the corporatization of civic technologies that would be heretofore publicly owned and controlled, and the prevailing paradigm of the smart city treats citizens as consumers and users of services, not as participants in the design of civic AI systems (Madaio & Martin, 2020). A long-standing issue with participation in general also applies to civic AI specifically, namely, superficial involvement lacking genuine influence (i.e., 'participation washing'). Furthermore, citizens are often framed as "customers" rather than active participants with a voice. Finally, there remains a need to develop concrete mechanisms for participation rather than an abstract recognition of its importance (Huang et al., 2025).

Reflecting on the small but growing body of work on civic AI, several tensions emerge: balancing technical AI knowledge with inclusive engagement, dominant AI systems' global, transnational nature versus local, community democracy, technical openness versus meaningful understanding for affected communities, and a gap between the current practice of public engagement which remains limited to education and awareness raising, and the aspirations of more fulsome participatory governance models.

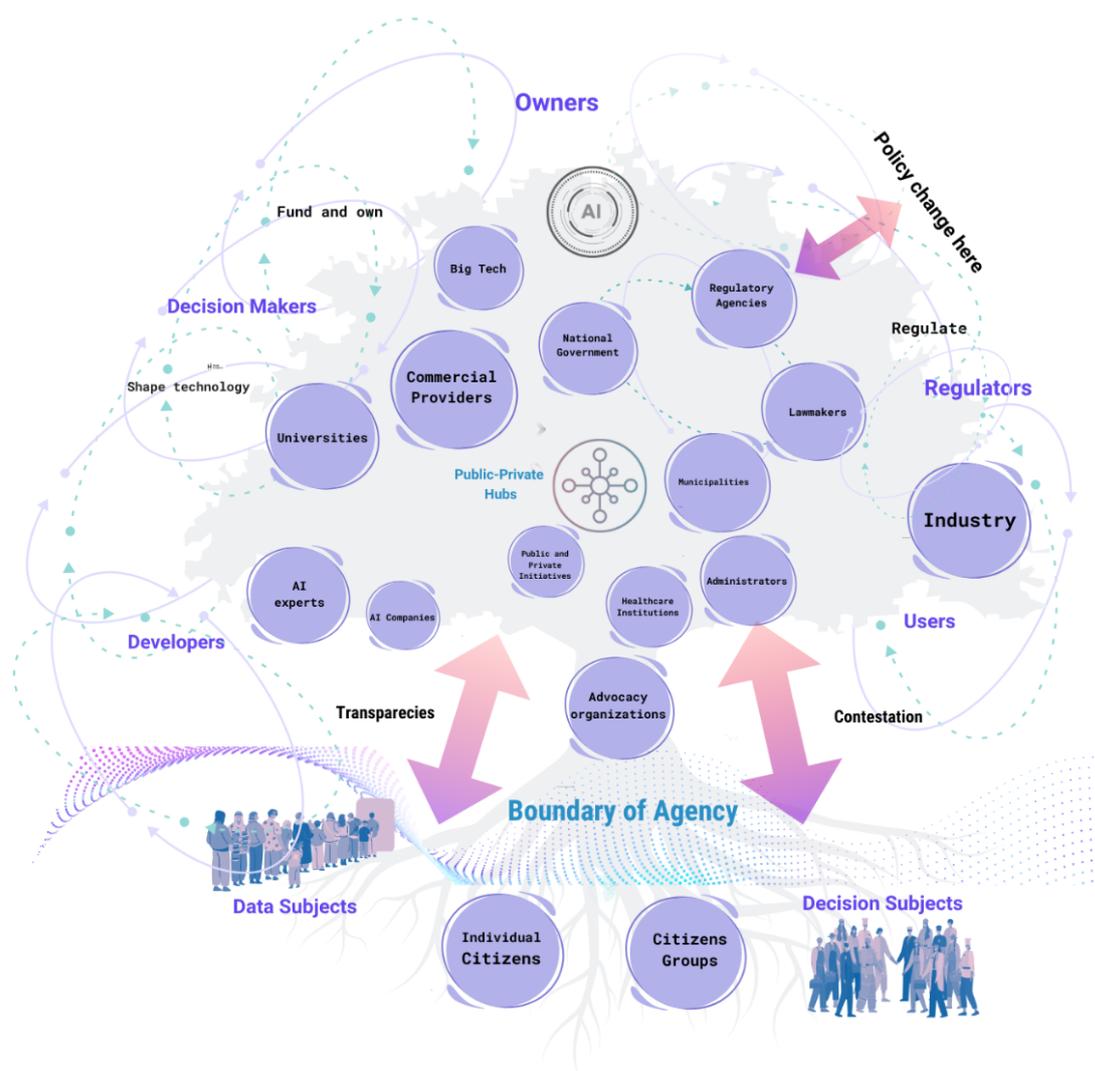

Figure 1: Overview of western European civic AI ecosystem, inspired by (Deshpande & Sharp, 2022), showing actors in the following rough groups: *owners* of technology fund and shape technology, with power and agency over systems' goals and scope; *decision makers* include the direct owners, public advisory boards, policymakers and external experts that shape goals, scope or operation of systems; *regulators* create the legal context of use of AI systems; *developers* create systems and components, which they may own or sell; *users* of the technology engage with the system outputs as part of their tasks, e.g. civic workers; *subjects* of technology are the people whose decisions are made about - *data subjects* are used to train and build models, and *decision subjects*, are subject of the AI systems decisions.

# 4 Conceptualising Meaningful Transparency in Civic AI

The challenges of Civic AI are particular, but not unique: we have complex overlapping systems contributing to decisions made about broadly defined groups of people, who are not always literate in the organisational and technical details of how decisions are made. Figure 1 illustrates the key stakeholders in a civic AI ecosystem, with reference Deshpande and Sharp's work (2022), in a form that should be fairly typical across OECD country contexts. The key takeaway is that AI systems development is a networked infrastructural activity connecting local and global companies, procurement practices, governmental organisations, policy, publics, decision subjects and so on.

In terms of the previous typologies (Section 2.1), we are primarily interested in: *downwards* and *inwards* transparencies, where decisions are made transparent to subjects, and to people outside the organisations (Heald, 2012; Meijer, 2014). A mixture of *events* and *processes* (Heald, 2012), but with a focus on the *linkage* processes from outputs to outcomes, and population-level outcomes. Primarily *operational* rather than *procedural* transparencies (Heald, 2012), as these relate to the specific cases of people's lives, although there are escalations from operational to procedural matters. A skew towards *real-time* rather than *retrospective* transparencies (Heald, 2012), as this is more likely to sit with the life-courses of decision subjects: the longer between the decision and the production of transparency, the harder it is to meaningfully engage, and the longer the period in which damage may be done. At the same time, the use of automated systems makes the production of real-time operational transparency increasingly plausible.

To deal with this complexity, we build on relational approaches proposed in (Christensen & Cheney, 2015; Felzmann et al., 2019; Meijer, 2014; van Nuenen et al., 2020; Van Nuenen et al., 2022; Weller, 2019), while also treating them situationally, as unique to particular contexts and groups of people and acknowledging that especially in the context of civic services, "transparency has limited consequences when the choice is nil." (Etzioni, 2016).

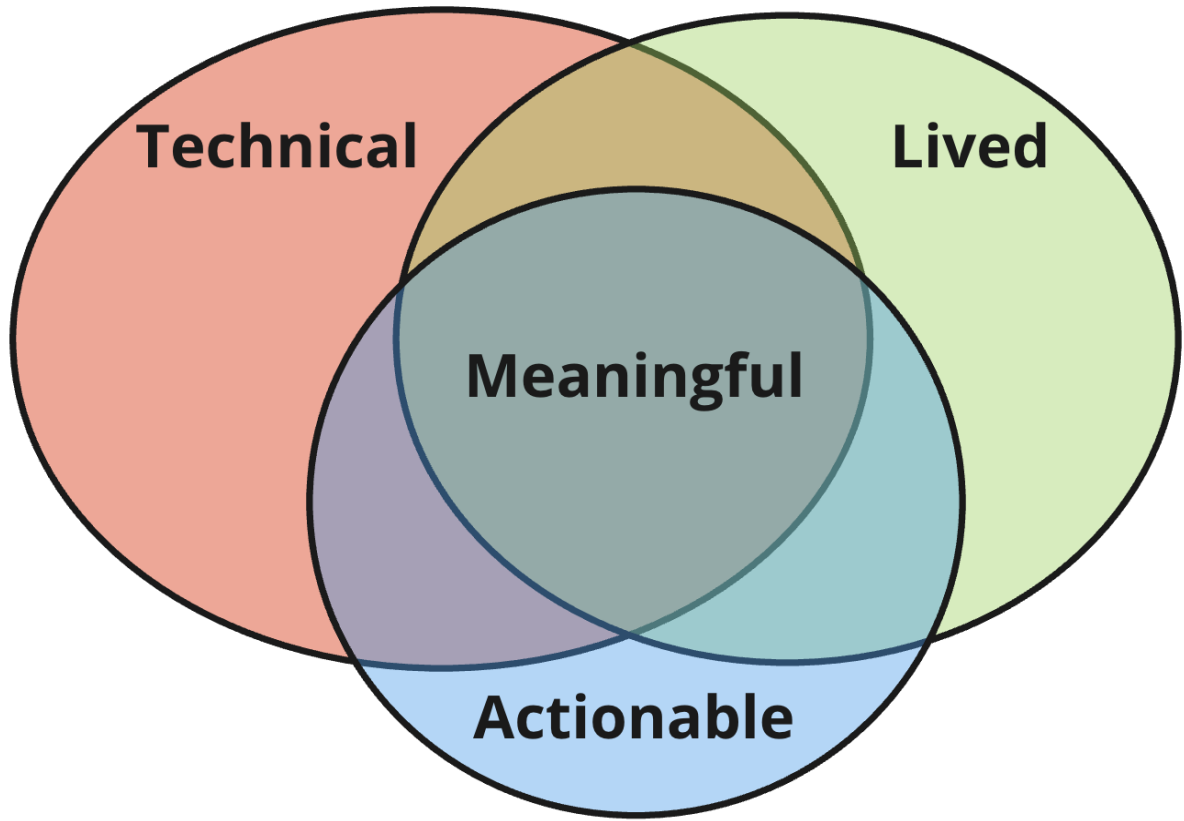

Figure 2: Construction of meaningful transparency for civic AI as the intersection of the lived transparencies that relate to residents' lives, are technically producible, and connected to the possibility for change and action

We suggest three overlapping qualities one might want from *meaningful* transparency (Figure 2):
- **Relates to lived experience**: Transparency should be a plural concept, with different transparencies for different people around different questions; it should cover qualities that are meaningful for the subjects of decisions (Abdu et al., 2024), without overloading them, and should be legible to a wide range of decision subjects, with a range of literacies and levels of engagement.
- **Actionable**: Transparency should be generative, and able to shape the organisation it renders visible (Albu & Flyverbom, 2019). Action may include production of accountability and direct contestation of decisions, but also the social sensemaking necessary to realise one is part of a group of people being discriminated against. The basis for action should extend beyond the analysis of particular model outputs or even processes, and support necessary engagement with the sociopolitical-material relations around the production of outcomes, including the commissioning and deployment of systems.
- **Technical production**: Transparency is only usable if it can be technically produced; so designing meaningful transparencies must take into account the potential to create them, along with safeguards on privacy, potentials for real-time production, cost to the service, and so on.

In the next sections, we present a discussion of the relation between transparency and lived experience for decision subjects, which is the critical novel component of this formulation of meaningful transparency. We then discuss the questions of actionability and technical production, as well as the intersections between these spaces.

## 4.1 Lived Transparency

The cornerstone of thinking into what would be meaningful in transparency for civic AI is a deep embedding in the lives of residents. We use 'lived transparency' to capture the idea of building part of the design and analysis of transparency systems from the ways that residents are experiencing and making meaning of the technical systems they interact with.

This approach builds on Husserl's idea of Lebenswelt/lifeworld/lived experience. Husserl asserted that human existence is fundamentally situational and contextual and must be studied from the viewpoint of the individual experiencing the phenomenon (Husserl, 1936). Schutz (1967) later connected Husserl's phenomenology to sociology, who argued that social interactions are integral to a shared Lebenswelt (lifeworld) of consciousness and subjective experiences. Across these works, "lifeworld" refers to the lived experiences of individuals within their social, cultural, and environmental contexts (Borkovich et al., 2024). Because it has become the more common term in recent works in HCI and AI, we henceforth use the term "lived experience" to refer to Lebenswelt/lifeworld.

Lived experience offers a way forward in the study of how humans experience information in context, sometimes called experienced (experiential) information (Bates, 2006; Poirier, 2012; Pollak, 2015). A benefit of phenomenologically inflected approaches is that we do not need to define experience, meaning, and context upfront. However, we can allow them to emerge throughout a study, as opposed to approaches that relegate these as secondary to observed action. When we work in this way, our findings do become less precise and less measurable, lacking in explanatory and predictive power. On the upside, our findings become more holistic and may uncover previously unseen things (Gorichanaz et al., 2018).

### 4.1.1 Lived Experience in AI Transparency Research

To illustrate the intersection of lived experience and AI transparency, we shortly describe four examples: i) Hassan and El-Ashry (2024) use interpretive phenomenological analysis (IPA) to "capture the lived experience" of critical care nurses in leadership roles working with AI. Instead of producing generalized findings, IPA offers insights into how a given person, in a given context, makes sense of a given situation. Their study shows how transparency is a significant factor in nurses' interpreting AI recommendations and, in turn, accepting (or trusting) AI. ii) Mager et al. (2025) propose "situated ethics" to ground global AI ethics in space and place. Situated ethics focuses on local contexts, concerns, and lived experiences. They aim to counter globalizing forces and anchor abstract values in the real world. Abstract notions of AI ethics are toothless when not grounded in real-world contexts and lived experiences. Transparency is one of the ethical principles they include as an example. iii) Nopas (2025) critically examines the role of AI-driven learning environments in shaping learner autonomy, knowledge co-construction, and digital learning equity in Thailand's online education landscape. They focus on the "lived experience of learners navigating AI-driven learning ecosystems." They find that a lack of transparency may lead to reduced learner autonomy. It should not control self-directed learning but only support it as a collaborative assistant. Greater transparency enables learners to make informed decisions about their "learning pathways."  iv) Lagerkvist et al.(2024) discuss the lived experience of biometric AI and how it objectifies "body, self, and identity." "Enforced transparency" is one of three ways this happens, referring to smart home assistants surveying human intimacies and "invading obscure spaces" like the bedroom. Note that this is a different kind of transparency from the one we focus on. In this case, people's private lives are made transparent to tech companies rather than the transparency of AI systems.

These four phenomenological uses of lived experience share a commitment to understanding how people experience AI systems in their specific social and material contexts, rather than relying on abstract principles or generalized assumptions about user needs.

Lived experience is more frequently invoked straightforwardly to refer to past or ongoing experiences of particular social or physical conditions that some groups of people share that are pertinent to the development or transparency of AI systems. For example, Lauderdale et al. (Lauderdale et al., 2025) argue for including more people with lived experience of mental disorders in GenAI development to reduce bias and increase accuracy before they are used in mental health applications. Guerrero Millan et al. (2024) invoke lived experience when discussing how participants view technology as extractive and invasive, which the authors suggest is connected to the participants' experiences with "authoritative practices." Anaduaka et al. (2025), working on the use of AI for predicting, identifying, diagnosing, and treating perinatal depression and anxiety (PDA) in low- and middle-income countries (LMICs), conduct focus groups that include "pregnant women and new mothers with lived experience of PDA." Crockett et al. (2024) work on integrating "co-production" methods into AI research so that "meaningful public engagement" is enabled. They highlight lived experiences influencing people's trust in AI in specific domains. "Lived experience of homelessness" is mentioned as one example of traditionally marginalized communities. Diversity and inclusion in co-production are linked to lived experience. The diversity they seek includes a diversity of lived experiences. This usage emphasizes lived experience as a form of expertise and knowledge that should inform AI development and transparency practices.

Lived experience is sometimes invoked to refer to factors that shape the positionality of authors themselves, such as when Lazem et al. (2022) use their own lived experience as outsiders to explore challenges in decolonizing HCI. This reflexive use acknowledges how researchers' embodied experiences shape their approach to studying technology and transparency.

Then, sometimes, lived experience is used to discuss how AI and computing artifacts do or do not capture, store, or represent the lived experiences of people. In the case of Martinez et al. (2025), working on GenAI in documentary photography, lived experience is invoked by participant photographers who doubt that AI can accurately represent lived experiences and fear AI could "compromise narrative integrity." In the case of Guerrero Millan et al. (2024), working on indigenous approaches to the development of technologies that support self-determination, lived experience is used when discussing prototypes that resulted from a co-design workshop to describe how participants see them as objects for storing past and present knowledge to pass on to future generations. These uses highlight tensions between technological representation and the rich, embodied nature of human experience.

*4.1.2 Towards Lived Transparency*
Our preliminary conceptualization of lived transparency emphasizes the situated, embodied, and contextual nature of how people experience and make sense of AI transparency. Unlike approaches that treat transparency as a universal good or a technical feature to be implemented, lived transparency recognizes that transparency's meaning and value emerge through people's concrete interactions with AI systems in specific contexts. This approach could deepen AI transparency research by revealing how transparency practices function in people's lives, uncovering misalignments between intended and experienced transparency, and highlighting the diversity of transparency needs across different communities and contexts. Rather than assuming what transparency should look like, lived transparency asks how transparency is lived.

## 4.2 Intersections between Lived, Technical, and Actionable Transparencies

### 4.2.1 Technically producible

We use *technical transparency* to cover transparency that is constructed through technical artefacts that *can* and *may* be produced with and around models or decision-making systems. The bulk of the work covered in Section 2.2 is approaches to producing technical artefacts and objects that support transparency—or explainability—in some way.

As a wide range of productions can support transparency in some way, it is easier to tackle the inverse question of what is *not* technically producible transparency.
- Despite the expanding frontier of XAI, there are many explanations or transparency features that would be desirable, but not currently possible. Some of these might change in the future, with better algorithms: for example, the desire for explanations of image categorisation prompted several approaches that shed light on what contributions in an image led to a classification (Balayn et al., 2021, 2022, 2023). In contrast, a precise account of which inputs or parameters in an LLM led to a particular word choice is unlikely to be forthcoming (although tools to spot output that matches training data are emerging (*OLMoTrace*, n.d.)).
- Some transparency artefacts may be technically producible, but social or legal factors prevent this. An example from outside of AI would be real-time financial trades, which could be produced, but are not legally required, as this would expose commercially sensitive information. Where public systems are concerned, privacy is a key factor, with sensitivity around personal data and the risks of differential disclosure contributing to reluctance to create particular views on decisions. Using tools and systems from external providers also means that commercial sensitivity is commonly invoked as a reason to avoid transparency.
- Transparency artefacts may be theoretically possible, but the organisation lacks the skills or resources to create them. Just as open data in government services has required significant effort, initiatives, and upskilling of workers, deploying transparency-capable AI systems will require the development of particular capacities. The form may be challenging—for instance, being able to look across a collection of cases to provide transparency around questions of bias requires additional development in both computational and statistical skills compared to handling individual processes.

### 4.2.2 Actionable

Actionable transparencies are ones that connect to the possibility of making change, whether through protest, contestation, democratic power, subversion, opting out, causing organisational change, and so on. Section 3.3 covered some of the reasons that transparencies do not support change; in contrast, reasons that a transparency is actionable would include:
- There is a formal mechanism that compels an organisation to change behaviour, whether through regulation, censure, legal means, or other force

- Organisational reputation is—or would be—damaged by the information being made transparent.
- Publics have some form of choice about the services they use, and can *exit* from ones that they find objectionable.
- Publics have some control over the way they are interpreted by these services (*voice*). This could include correcting incorrect data, selecting alternative processing techniques, or altering behaviour to comply with system requirements. In classification systems, the idea that one can change behaviour to change ones classification is labelled *actionable resource,* and support by techniques such as generating counterfactual examples that illustrate the desired classification to indicate which variables the subject might be able to affect (Rasouli & Chieh Yu, 2024; Ustun et al., 2019) .
- Publics can productively engage with the institutions providing services to shape or co-design the way that they operate.

Particularly in the civic context, where services are provided on behalf of the population, there is an expectation that transparency could and should lead to positive change. It is reasonable to think organisations are well-intentioned with respect to the population, and a combination of transparency and citizen engagement is a route towards better provision.  A fuller treatment of ways to ensure transparencies are actionable is given in Section 6, in particular through participation, contestation, and agonism.

## 4.3 Constructing Meaningful Transparency

Based on this, we can ask of any conceived transparency mechanism: is it technically producible? Does it connect to lived experience? Does it support the potential for action? When a transparency hits all three areas, we term it *meaningful*: it supports citizens making change about something important to them. Many pieces of work can be situated at the intersections of these concepts: disclosure interfaces (Norval et al., 2022) use HCI concepts to relate technical outputs to particular audiences; infrastructuring projects that work against atomisation (L. C. Irani & Silberman, 2013) connect lived experience to the potential for action; algorithm registries and related projects (Schuitemaker et al., 2024) provide substrates to connect technical outputs to meaningful change.

It is worth noting that these questions are all situational and relational: they are not answered for a general mechanism, but rather for specific techniques in particular contexts. To take a non-AI example, a Freedom of Information (FoI) request is a tool for producing transparency about civil organisations. For many individual citizens, it is not a lived transparency, as the mechanism and results are opaque; for committed citizen groups, however, it can be a useful tool for them to directly acquire and interpret information that they care about—the process of group action brings it from a technical process into one that can be connected to lived issues. The context around a transparency production can change whether it is technically producible through socio-legal restrictions or the resources of the organisations involved; it can change whether it is actionable, through the regulations around the technology, the possibilities of complaint, contestation, bad publicity forcing organisational change and so on; and the particular affected population will have a determining effect on whether it is lived. The production of transparency for clinicians (B. G. S. Schor et al., 2024b) is *meaningful* for the clinician: it speaks in their language, of features they are sensitive to, and supports their decision making; it is less *meaningful* to the patients, except indirectly through the provision of better medical care.

The possibility of realising meaningful transparency is hence built on techniques to engage diverse stakeholders, working with a sensitivity to their particular needs, with a strong connection to the complexities of the particular socio-technical context that they are embedded in - all while maintaining technical rigour and connection to the organisational and political levers that enable change.

# 5 Discussion: Emerging Practices to Enact Meaningful Transparency

In the previous section, we detailed three desirable qualities that comprise *meaningful transparency*: that it is actionable, that it can be technically produced, and that it connects to people's lived experiences. Here, we survey three key practices that can support these goals, and help to ensure that systems are meaningfully transparent from their conception through their design and deployment: participation, co-design and co-sensemaking; contestation and agonism; and transdisciplinary methodologies.

## 5.1 Participation

Involving the public in the development of AI systems fosters a shared understanding of the technology (Parker et al., 2025), enhances the agency of decision-subjects (Corbett et al., 2023), and creates opportunities for contesting design decisions (Alfrink, Keller, Doorn, et al., 2022). As such, participation can be a way to promote, enact and prefigure meaningful transparency in civic AI. In this section, we will discuss current and emerging participatory AI practices in relation meaningful transparency.

*5.1.1 Participatory AI*

There is an increasing demand across various disciplines and sectors to involve the public in the development of AI systems for public organizations to increase transparency (Birhane et al., 2022b). Although the approaches differ both epistemologically and methodologically, the call for participatory practices is widespread (Delgado et al., 2023; Zytko et al., 2022).

Participatory processes, such as participatory design (Bannon & Ehn, 2012), participatory action research, and deliberation methods, involve recognizing that not only engineers and designers can contribute to the design of technology (N. B. Hansen et al., 2020). In this framework, individuals labeled as "non-experts" and users possess valuable knowledge based on two key aspects: (i) their lived experiences with technology and (ii) their unique experiential understanding and expertise related to the specific subject matter or societal challenge that the technology aims to address. This knowledge is closely linked to their relational, cultural, and spiritual insights (Udoewa, 2022). Overall, participatory practices aim to democratize design by involving "end-users" as experts in their experiences, valuing their lived experience in context, and establishing new forms of collaborations between designers and "end-users" (Björgvinsson et al., 2010).

Participatory practices take many forms. They can shape decisions and actions through building aggregate perspectives using qualitative methods such as focus groups (Abebe & Goldner, 2018), or developing methods for democratic decision-making, political action, and public engagement (OECD, 2020). Approaches such as *co-design* are used to harness people's sense-making and creativity, allowing them to contribute to the design of technology that affects them (Sanders & Stappers, 2008; Steen, 2013). To achieve this, specific participatory activities and co-design toolkits are developed that enable individuals who are not trained in design to collaborate in the generation of tangible ideas (Sanders & Stappers, 2014). In participatory action research, stakeholders become co-inquirers to co-construct research plans and interventions with researchers (Cornish et al., 2023).

Participatory AI stems from the premises of these participatory practices (Corbett et al., 2023; Zhang et al., 2023). The participatory turn, as described by Birhane et al. (2022b), relates to acknowledging "that the communities and publics beyond technical designers have knowledge, expertise and interests that are essential to the development of AI that aims to strengthen justice and prosperity" (p. 8). Take a particular example, *epistemic injustice* is wronging an individual in their capacity as a knower (Fricker, 2007), discounting their experience because of who they are (testimonial injustice) or because their experience is not recognized as valid (hermeneutic injustice). AI systems have been shown to negatively impact epistemic justice, by encoding bias about individuals identity in the decision-making process, by erasing their lived experience in system outputs and so on(Kay et al., 2025; Laacke, 2023; McInerney, 2025; Pohlhaus, 2017). Arguably, by including the decision subjects and the public in the process, they could counteract these harmful effects, increasing their agency in determining how AI works and impacts their lived experience, offering an opportunity to actively contest and imagine alternatives (Alfrink, Keller, Kortuem, et al., 2022a).

*5.1.2 Participation and Meaningful Transparency in Civic AI?*
Participatory AI practices have a unique ability to promote meaningful transparency. By emphasizing lived transparency—the understanding of how individuals experience and interpret AI transparency in a specific context—these practices connect with technical transparency. This connection fosters a collaborative environment where stakeholders and developers can engage in AI development that pays attention to public epistemologies while being mindful of the power dynamics, legal and social relationships that influence the development and deployment of the systems.

In particular, we here provide some illustrative examples of how participation in the design of civic AI may enact meaningful transparency in three different frames: co-design of systems as they come into being; co-sensemaking of what systems are doing in practice; and control or contestation as the system operates.

*Co-design*: bringing together citizens, civil society, domain experts, and developers collaboratively ideate and critique AI solutions helps develop socially situated approaches to transparency. For example, in Finland, workshops were held to gather residents' perspectives on what a "trustworthy public-sector AI service" should look like (Drobotowicz, 2020). Such workshops use plain language and scenarios to prompt discussion—e.g., "How would you feel about an algorithm deciding your eligibility for a subsidy? What would you need to trust it?". The outcomes inform design guidelines – the guidelines for public sector AI from one Finnish study were directly shaped by citizen input, highlighting needs like data consent options and human review (Drobotowicz, 2020).

*Co-sensemaking*: collective action to establish shared understanding of how systems work helps build meaningful ideas about what transparencies are needed. For example, participatory workshops have been used to identify what information marginalized communities want from ADM systems in welfare or housing decisions (Drobotowicz et al.,

2023). Such approaches recognize lived experience as a form of expertise—the people who continually deal with an algorithm often know best where it fails or confuses. One study argues that "everyday users of algorithmic systems are often best poised to detect and raise awareness about harmful behaviors" they encounter (Shen et al., 2021; Van Nuenen et al., 2022). By involving these users in transparency efforts—through feedback loops, user councils, or citizen audits—designers can co-create explanations or disclosure practices that address real concerns. An example of co-construction in practice is the development of community guidelines for algorithmic systems, where affected users help draft the explanation templates or fairness criteria that the system should adhere to. This relational, participatory ethos is sometimes contrasted with the top-down "expert-driven" transparency of early AI governance.

*Interactional Participation*: a well-designed civic AI can give agency to individual users at the moment of interaction. For example, a digital platform for an AI-driven welfare system might allow the user to choose their level of involvement—e.g., "Do you want to provide additional data for a potentially more personalized decision, or proceed with minimal data?"—or even let them opt out of the algorithm's assistance in favor of a human officer (Drobotowicz, 2020). In a thesis project on "trustworthy AI services," Drobotowicz (2020) proposed that citizens should be able to toggle the AI's role – from fully automated to advisory and revert or disable the AI if they are not comfortable (Drobotowicz, 2020). Concretely, during a consent stage, the system could ask the citizen: "Do you consent to this AI making an initial decision on your case? You can opt out and have a human process it instead (which might take longer)." Providing such choices can dramatically increase users' sense of control. Furthermore, user agency is enhanced by negotiability—the ability for users to question the AI or provide corrections. Imagine an AI that explains, "We found you ineligible based on X and Y," and the user can respond, "Actually, Y is incorrect in my data," prompting the system to re-evaluate - connecting the technical transparency about the decision making to a potential user action. Some advanced interfaces and "explainable AI" research are moving toward this kind of interactive explanation and correction loop, essentially a dialogue between citizen and AI rather than one-way decisions (Gensinger, 2024).

These forms of participation are not exhaustive, but paint a picture of emerging possibilities to use participatory practices in civic AI systems to enhance meaningful transparency.

## 5.2 Contestation, Agonism and Beyond
Another way to enact or prefigure meaningful transparency is through contestability.

When transparency is challenging, Walmsley (2021) argues that contestability offers a viable alternative. Decisions can be contested without fully understanding how they were made. This may involve civic participation during development, human oversight during decision-making, or feedback from affected subjects to system developers. Even without complete understanding of outputs, we can still challenge them (Walmsley, 2021).

Contestable AI seeks to safeguard against flawed and unjust machine intelligence by emphasizing human involvement and fostering adversarial debate between those subjected to AI systems and those designing, developing, and deploying them (Alfrink, Keller, Kortuem, et al., 2022b; Almada, 2019a; Capel & Brereton, 2023; Eslami et al., 2019; Henin & Le Métayer, 2022; Hirsch et al., 2017; Sarra, 2020). Contestability can be viewed as humans questioning machine predictions, allowing human intervention to correct potential machine errors (Hirsch et al., 2017; Verma et al., 2023). It can be described as a mixture of human and machine decision-making, emphasizing its role in procedural justice and enhancing perceived fairness (Eslami et al., 2019; Lyons et al., 2023; Yurrita et al., 2023). The practice of "contestability by design" stresses human intervention in AI development processes (Almada, 2019b). Contestability transcends simple human intervention, demanding a dialectical interaction between decision subjects and human controllers (Sarra, 2020). Contestability is the mechanism by which people can demand justifications (i.e., normative accounts) in addition to explanations (i.e., factual accounts), hold system operators accountable, and, in turn, legitimate a system (Henin & Le Métayer, 2022). Contestable AI can be conceptualized as systems that are open and responsive to dispute, throughout their lifecycle, emphasizing a dialogical relationship between system operators and subjects (Alfrink et al., 2023, 2024; Alfrink, Keller, Kortuem, et al., 2022b).

The political ideal that underwrites the project of contestable AI is in keeping with agonistic pluralism (Mouffe, 2007)[1]. Agonists argue for a "return of the political" (Mouffe, 1995)[2] and for perceiving contestation as something to be celebrated. Agonistic pluralism is characterized by a commitment to radical pluralism (a diversity of values is constructive, not needing resolution), a tragic view of the world (where conflict is ineradicable and intrinsic to social relations), and a conviction that conflict can be productive (Lowndes & Paxton, 2018, pp. 694–696).

A generative metaphor that can guide the design of AI systems for contestability is the "Agonistic Arena" (Alfrink et al., 2024)[3]. Agonistic pluralism is the political philosophy underpinning this metaphor. For the image of the arena, this metaphor takes inspiration from the ancient Greek ideal of democratic competitiveness (Filonik, 2022). This metaphor casts AI systems as a space in which conflict in various forms is embraced and celebrated as a productive force. It finds expression as practices that seek to establish new discursive relations between stakeholders, enable continuous monitoring in the interest of contingency and admittance of fallibility, and create sociotechnical arrangements prioritizing mutability and reversibility.[4]

---

[1] Mouffe's agonistic pluralism views democratic politics as necessarily involving conflict and the us/them distinction, but seeks to transform antagonistic relations between enemies into agonistic relations between legitimate adversaries who share democratic principles while disagreeing on their interpretation, recognizing that consensus without exclusion is impossible and that power relations are constitutive of all social arrangements. For a succinct summary of this approach to democracy, see Mouffe (2007, pp. 42–44).

[2] According to Mouffe, "the political" refers to the dimension of antagonism that exists potentially in human relations, while "politics" seeks to establish a certain order and organize human coexistence in conditions that are permanently conflictual because they are affected by "the political." For a brief discussion of the distinction between politics and the political, see Mouffe (1995, pp. 105–106).

[3] Generative metaphors are guiding concepts that influence perception and understanding of the world. Generative metaphors function by transferring perspectives between domains. The resulting perception affects decisions and actions, including the ways in which designers frame and solve problems (Schön, 1993).

[4] For an overview of how agonistic pluralism has informed work in human-centered AI more broadly, and a more extensive description of this metaphor, see Alfrink et al. (2024, pp. 58–59).

Contestable AI enacts or prefigures meaningful transparency across its three dimensions. Technically, it provides concrete mechanisms, such as appeals processes and tools for external oversight, that can be implemented to increase transparency. With regards to lived transparency, it connects transparency to people's lived experience of AI systems by giving them the ability to interrogate and intervene in systems when something seems wrong. And as actionable transparency, it emphasizes the connection between information and means of control and seeks to create ways for people to intervene beyond passively observing a system.

Contestability overlaps but also differs the participatory, and co-sense-making approaches discussed in the preceding.

With co-sensemaking and epistemic justice, it shares the emphasis on valuing the knowledge that comes from the lived experience of communities and challenging expert-dominated narratives. However, contestability focuses more specifically on dispute mechanisms rather than broader forms of knowledge co-construction.

With participatory approaches, contestability, particularly in the form of contestability by design, shares a desire for the further democratization of AI systems development and governance. Both would argue that communities and publics beyond technical designers have knowledge, expertise, and interests that are essential to the development of AI. Where contestability differs is that its emphasis is on structured adversarial processes rather than collaborative consensus-building. Where participatory approaches can face challenges of co-optation and power asymmetries, contestability's agonistic framing explicitly anticipates and works with conflict rather than seeking to resolve it into consensus.

By embracing structured conflict rather than consensus-seeking, contestability offers a distinctive pathway to meaningful transparency that acknowledges the inherently political nature of AI deployment in civic contexts. This agonistic approach may prove particularly valuable where power asymmetries and diverse stakeholder interests make collaborative consensus-building insufficient for genuine democratic accountability.

## 5.3 Transdisciplinarity and Meaningful Transparency

Governments in the EU and beyond are beginning to realize that algorithmic governance requires democratic governance—you cannot have machines governing people without people governing the machines. As a result, transparency, consultation, and participatory design are becoming standard recommendations for any public sector AI project (Fischer, 2021; Merchant, 2023). While true power-sharing with citizens in AI governance is still nascent, there is a clear trend toward more openness. The participatory practices detailed above are used to increase agency and epistemic justice in Civic AI in two directions: *upstream* practices include co-sense-making, co-designing, and collective decisions about what and whether to build; *downstream* practices include ensuring users can meaningfully interact with or contest AI systems, while having enough room to imagine alternatives. However, there are several limitations to the scope and effectiveness of participation (Ayling & Chapman, 2022; Birhane et al., 2022b; Sloane et al., 2022; Suresh et al., 2024).

- While there is considerable discussion about participation, citizens often lack genuine decision-making power. Sieber et al. (2025) conclude that, to make public participation in AI truly meaningful, it is essential to address power imbalances and move beyond "neoliberal intent," which refers to tokenistic or market-driven approaches to participation. In other words, merely holding a focus group is insufficient; the process must be thoughtfully designed to genuinely value and incorporate citizens' input, even if this requires a change in project direction.
- The focus on reaching consensus in participatory sessions causes a flattening of diverse values and worldviews, erasing the productive conflict inherent to relational, agonistic approaches (Geppert & Forlano, 2022).
- While one of the goals of participatory practices is to address power structures, practical and effective methods to address power differences among stakeholders and designers are currently lacking (C. Harrington et al., 2019). Designers' good intentions in participatory practices often lead to the unintentional reproduction of privileged perspectives (C. N. Harrington, 2020; C. N. Harrington et al., 2024); By breaking promises of democratization, participation often ends up prioritizing corporate or political interests over social justice.
- The technical complexity of AI systems means it is hard to resolve the asymmetry of knowledge between designers and the people. This can be partially addressed through public education on ways to understand and make sense of AI systems (Blythe et al., 2025; Lupetti & Murray-Rust, 2024; Murray-Rust, Nicenboim, et al., 2022), as with other technical infrastructures (Murray-Rust, Elsden, et al., 2022). It may also require new intermediaries, such as the data intermediaries proposed in the UK, to help manage data sharing and empower individuals (Merchant, 2023).
- Finally, making participation truly representative is extremely difficult: often, the voices heard are those of already empowered groups, so special effort is needed to include marginalized communities, potentially through partnership with community organizations.

In the face of this, enacting meaningful transparency may require an epistemological shift towards *transdisciplinary* research practices that transcend disciplinary boundaries and that value lived experience as a source of knowledge. Transdisciplinary research (Pohl & Hadorn, 2007) is a socially engaged way to practice knowledge production that is increasingly explored in design research from a variety of perspectives (Ozkaramanli et al., 2022; Thompson Klein, 2004; van der Bijl-Brouwer, 2022). The premise of engaging in transdisciplinarity is that complex societal challenges cannot be understood or tackled by a single academic discipline (Nicolescu & Ertas, 2008) or a particular group or community in isolation (Mobjörk, 2010). True transdisciplinarity involves questioning the existing norms of scientific knowledge and fostering new forms of collaboration among scientists, designers, governments, companies, and the public. The goal is to develop integrated knowledge and facilitate social transformations in public organizations through personal and collective learning. Transdisciplinary knowledge emphasizes relationships rather than individual components, and views the interactions between academia and civil society as essential for addressing complex, socially embedded problems. Knowledge integration across heterogeneous stakeholders demands facilitation and conflict mediation, alongside reflexivity and an awareness of power dynamics.

Transdisciplinary research methods often conflict with conventional knowledge production approaches in AI research. Transdisciplinary methods are characterized by responsiveness over top-down planning, context-specific nature, and emergent qualities. This emphasis on situational and integrated knowledge does not easily align with traditional ways of assessing rigor and validity around discrete units of carefully bounded work. In addition, detailing, practicing, and documenting transdisciplinary work is a challenging process: to capture the complexity of the process, to notice and record the unfolding dynamics and moments of knowledge production, to gather the breadth of threads necessary to understanding the various layers at play is a subtle and demanding task.

As a result, there is still much to be defined regarding how to generate transdisciplinary knowledge and address issues of values, pluralism, participation, and the integration of expertise in the socio-politically complex environment of deploying advanced computational systems. While transdisciplinarity is an appealing epistemological shift, significant work remains ahead.

If transparency is to be made meaningful for publics, the practice of constructing it *with* them, going to the heart of their actual concerns and lived experiences around AI systems, entangling with their diverse viewpoints and plural conceptions, entrenched positions, particular skills, understandings and literacies seems to be a necessary step. Civic AI should not be designed simply *for* people, but *with* people, to ensure it genuinely serves the public interest and supports meaningful transparency. To do so, we need to shift to more socially oriented forms of knowledge production.

# 6 Concluding remarks

In this paper, we have set out a working definition for the concept of *meaningful transparency* in Civic AI systems. This is built on a combination of digging into the foundations of transparency, and in particular its construction as a relational concept within a socio-technical understanding of AI systems as complex networked infrastructures. We have presented a discussion at the intersection of transparency and lived experience to construct what we term *lived transparencies*, and brought this together with the *technical* production and *actionability* of transparencies. This casts *meaningful transparency* in civic AI systems as transparencies that can be technically produced, connect to the lives of citizens, and that support the possibility for action. This sets a stage for future work, based on participation, contestation and agonism and the adoption of transdisciplinary practices as ways to ensure that civic AI systems are built with the kinds of transparency that matters to the publics they affect